\pdfoutput=1

\documentclass[11pt]{article}

\usepackage[table,xcdraw]{xcolor}
\usepackage[final]{acl}

\usepackage{multirow}
\usepackage{times}
\usepackage{tabularx}
\usepackage{latexsym}
\usepackage{graphicx}
\usepackage{array}
\usepackage{arydshln}
\usepackage{makecell}
\usepackage{booktabs}
\usepackage{siunitx}
\usepackage{caption}
\usepackage{CJKutf8}

\definecolor{ggreen}{HTML}{d4e7cf}
\definecolor{bblue}{HTML}{00bfff}
\definecolor{ppink}{HTML}{ff69b4}
\definecolor{yyellow}{HTML}{fffaa0}
\usepackage[T1]{fontenc}

\usepackage[utf8]{inputenc}

\usepackage{microtype}

\usepackage{inconsolata}
\usepackage{caption}
\usepackage{subcaption}

\title{Cross-lingual Back-Parsing: Utterance Synthesis from Meaning Representation for Zero-Resource Semantic Parsing
}

\author{
Deokhyung Kang$^{\spadesuit}$,
Seonjeong Hwang$^{\spadesuit}$,
Yunsu Kim$^{\heartsuit}$,
Gary Geunbae Lee$^{\spadesuit\diamondsuit}$\\
\normalsize{$^\spadesuit$Graduate School of Artificial Intelligence, POSTECH, Republic of Korea} \\
\normalsize{$^\diamondsuit$Department of Computer Science and Engineering, POSTECH, Republic of Korea} \\
\normalsize{$^\heartsuit$aiXplain, Inc. Los Gatos, CA, USA} \\
\normalsize{\texttt{\{deokhk, seonjeongh, gblee\}@postech.ac.kr}, \texttt{yunsu.kim@aixplain.com}}
}

\begin{document}
\maketitle
\begin{abstract}
Recent efforts have aimed to utilize multilingual pretrained language models (mPLMs) to extend semantic parsing (SP) across multiple languages without requiring extensive annotations. 
However, achieving zero-shot cross-lingual transfer for SP remains challenging, leading to a performance gap between source and target languages.
In this study, we propose \textbf{C}ross-lingual \textbf{B}ack-\textbf{P}arsing (\textbf{CBP}), a novel data augmentation methodology designed to enhance cross-lingual transfer for SP. 
Leveraging the representation geometry of the mPLMs, CBP synthesizes target language utterances from source meaning representations. 
Our methodology effectively performs cross-lingual data augmentation in challenging zero-resource settings, by utilizing only labeled data in the source language and monolingual corpora.
Extensive experiments on two cross-lingual SP benchmarks (Mschema2QA and Xspider) demonstrate that CBP brings substantial gains in the target language.
Further analysis of the synthesized utterances shows that our method successfully generates target language utterances with high slot value alignment rates while preserving semantic integrity.\footnote{Our codes and data are publicly available at \url{https://github.com/deokhk/CBP}.}

\end{abstract}

\section{Introduction}
Semantic Parsing (SP) is the task of converting natural language utterances into meaning representations such as SQL or Python code. With numerous English parsing datasets available, recent studies have enabled applications ranging from natural language interfaces for databases to code generation ~\citep{le2022coderl,li2023resdsql}. Despite SP's practicality, extending it beyond English is challenging. Manually annotating examples for other languages is very costly, and relying on machine translation is often impractical due to the complex slot alignment step after translation~\cite{nicosia2021translate}.

Recent studies focus on leveraging multilingual pretrained language models (mPLMs)~\citep{devlin-etal-2019-bert, xue2021mt5} to extend SP across multiple languages without costly annotations~\citep{sherborne2022zero, held-etal-2023-damp}. 
After being pretrained on large-scale non-parallel multilingual corpora, mPLMs demonstrate strong zero-shot cross-lingual transferability: 
Once these models are fine-tuned with labeled data from the source language, they show remarkable performance in target languages without using any labeled data from the target language.
Nonetheless, zero-shot cross-lingual transfer for SP is still challenging for state-of-the-art multilingual models, resulting in a notable performance gap between the source and target languages~\cite{ruder2021xtreme}.

\begin{figure}[!t]
    \centering
    \includegraphics[width=\columnwidth]{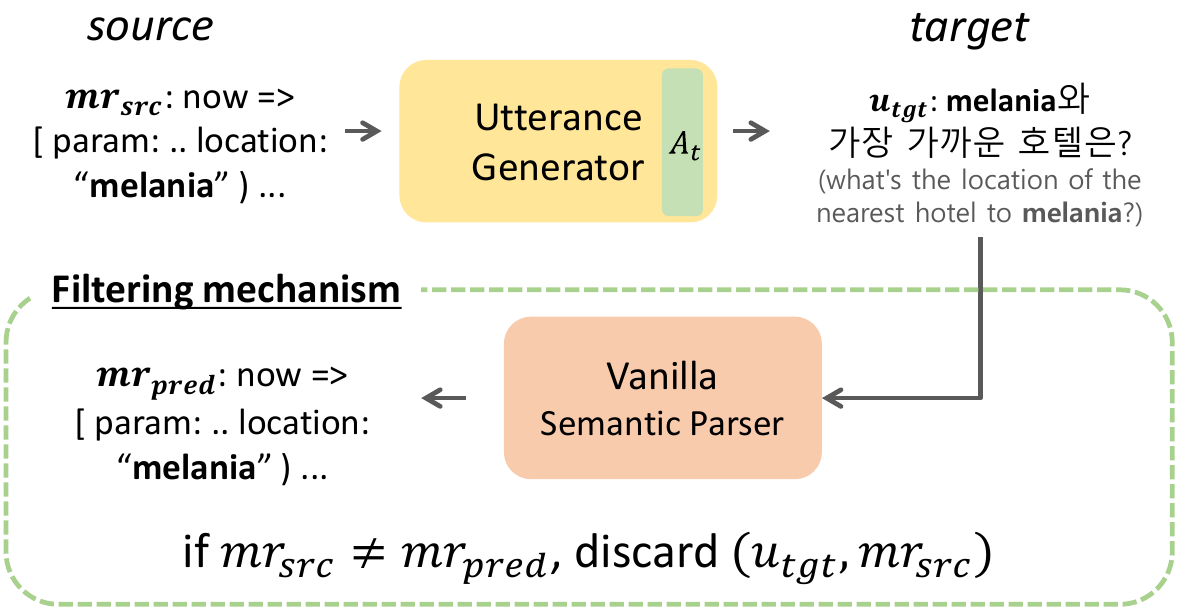}
    \caption{An overview of the data augmentation process with CBP. The utterance generator equipped with the language adapter $A_{t}$ synthesizes utterances in the target language $t$, and a filtering mechanism is applied to discard low-quality utterances. In the example, a Korean utterance is generated, with the corresponding English translation provided in parentheses.}
\label{fig:overview}
\end{figure}

To this end, we propose \textbf{C}ross-lingual \textbf{B}ack-\textbf{P}arsing (\textbf{CBP}), a novel data augmentation methodology for enhancing zero-shot cross-lingual transfer for SP. CBP is designed to be widely applicable by synthesizing target utterances from source meaning representations under zero-resource settings - where resources such as translators, annotated examples, and parallel corpora in target languages are unavailable. As shown in Figure~\ref{fig:overview}, CBP comprises two components: an utterance generator synthesizing utterances in the target languages and a filtering mechanism discarding low-quality utterances. 

To synthesize target utterances in the zero-resource setting, the utterance generator leverages a multilingual pretrained sequence-to-sequence (seq2seq) model such as mT5~\cite{xue2021mt5} with modular language-specific adapters~\cite{houlsby2019parameter,pfeiffer2020mad} inserted into the decoder. 
To enable the model to generate output text different from the input, we design a novel\emph{ source-switched denoising objective} for training the language adapters, leveraging findings~\cite{yang2021simple} that the language identity component can be extracted from contextualized representations of mPLMs. 
Using unlabeled target language sentences, we train the adapters to denoise input sentences from encoded representations with their language identity switched to the source language. This allows the adapters to control the output language of the utterance generator during inference.

We then synthesize target utterances from the source meaning representations using the utterance generator equipped with the target language adapters. This process effectively performs data synthesis to create new target language utterances, serving as data augmentation. Finally, we filter these synthesized utterances to discard low-quality ones using a filtering mechanism inspired by round-trip consistency~\cite{alberti2019synthetic}, thereby enhancing the quality of the augmented dataset.

We assess the efficacy and robustness of CBP on two challenging cross-lingual SP benchmarks, Mschema2QA~\cite{zhang2023xsemplr} and Xspider~\cite{shi2022xricl}, encompassing a total of 11 languages. 
In the Mschema2QA benchmark, CBP notably improves the average exact match by 3.2 points. 
Utilizing solely monolingual corpora for data augmentation, CBP surpasses all baselines that rely on translator-based data augmentation. 
For the Xspider benchmark, CBP exceeds the state-of-the-art, improving the exact match for Chinese from 52.7 to 54.0. 
Extensive analyses substantiate the effectiveness of our methodology. 
Further investigations into synthesized utterances indicate that CBP successfully generates utterances in the target languages high slot value alignment rates while moderately preserving semantic integrity, despite the absence of parallel corpora.

\begin{figure*}[!ht]
  \begin{subfigure}[b]{0.46\textwidth} 
    \includegraphics[width=\textwidth]{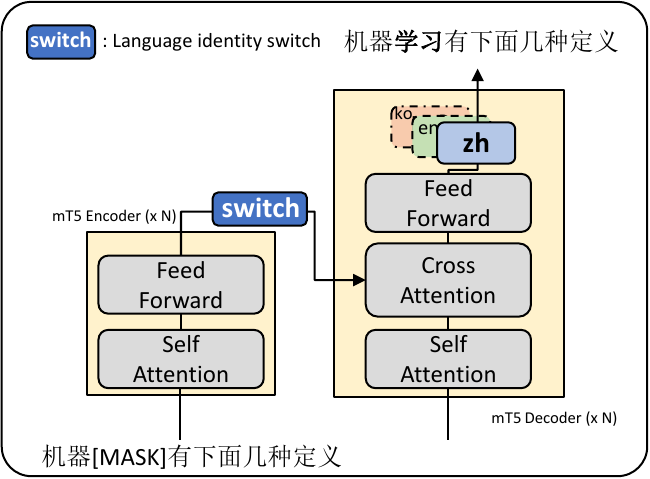}
    \centering
    \caption{Step 1: Language adapter training with \textbf{monolingual data}}
    \label{fig:step1}
  \end{subfigure} \hspace{0.8cm}
  \begin{subfigure}[b]{0.46\textwidth} 
    \includegraphics[width=\textwidth]{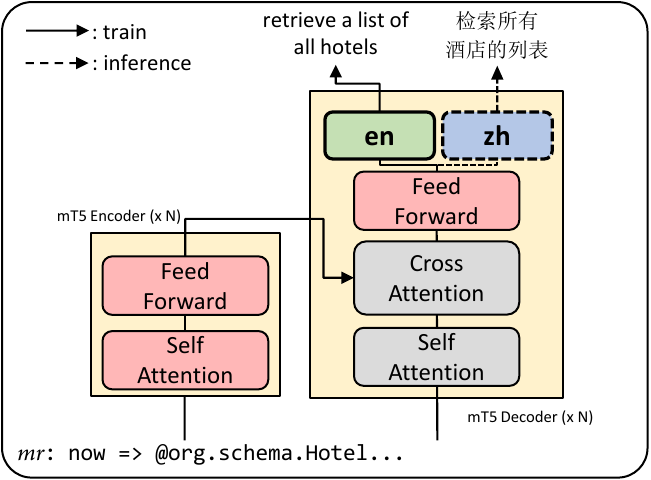}
    \caption{Step 2: utterance generation training and inference with \textbf{labeled data}}
    \label{fig:step2}
  \end{subfigure} 
  \caption{Overview of the utterance generator training. First, we individually train language adapters, represented by colored boxes in the decoder, using monolingual corpora for each language through source-switched denoising training. The remaining shared parameters are frozen during this process (\ref{fig:step1}). Next, we use labeled data to fine-tune the utterance generator for the utterance generation task, keeping the trained adapters frozen while selectively training the other parameters (\ref{fig:step2}). The figure is adapted from \citet{ustun2021multilingual}.}
  \label{fig:ZeroAug}
\end{figure*}

\section{Related Work}

\paragraph{Zero-shot cross-lingual semantic parsing} 
Zero-shot cross-lingual SP aims to transfer parsing capabilities from a high-resource language (e.g., English) to low-resource languages without requiring any training data in the low-resource languages. 
To enhance cross-lingual transfer, several studies introduce auxiliary objectives during training to improve the alignment of semantic spaces between languages~\cite{sherborne2022zero, held2023damp}. 
Our method, however, aligns with a different line of research: data augmentation. 
~\citet{xia2021multilingual} utilize machine translation to convert English datasets into target languages, followed by word aligners to match corresponding elements, whereas ~\citet{nicosia2021translate} directly generate aligned datasets using a fine-tuned model. 
Although not in the zero-shot setting, some works prompt large language models (LLMs) to generate synthetic data in the target language, using a few examples in the target language \cite{rosenbaum2022clasp, awasthi2023bootstrapping}. 
In contrast, our research addresses data augmentation in a relatively unexplored zero-resource setting, where no target language data, translators, or parallel corpora are available. Our approach leverages multilingual pretrained language models and monolingual corpora in the target language for augmentation, ensuring effective cross-lingual transfer without such resources. 

\paragraph{Multilingual language models}
Research on the representation geometry of multilingual pretrained language models (mPLMs) has revealed that the encoder representations of these models possess a shared multilingual representation space while still encoding language-specific information. A study by~\citet{libovicky2019language} shows that subtracting the language mean from representations enhances cross-lingual transfer by inducing language-agnostic representations. Additionally,~\citet{chang2022geometry} demonstrates that projecting representations onto language-specific subspaces can facilitate token predictions in specific languages. Leveraging these findings,~\citet{yang2021simple} enhances cross-lingual retrieval performance by removing language information from multilingual representations, while~\citet{deb-etal-2023-zero} improves cross-lingual question answering by projecting source representations onto target language subspaces during fine-tuning. The study most closely related to ours is by~\citet{wu2022laft}, which enhances cross-lingual transfer in natural language generation tasks by selectively removing language identity during the training process of a multilingual seq2seq model. In contrast, our approach leverages the model's language identity to modify the generation language of an already fine-tuned multilingual seq2seq model, enabling cross-lingual data augmentation in a zero-resource setting.

\section{Methodology}
\paragraph{Overview} In this study, we synthesize target language utterances \textit{u}$_{tgt}$ from source language meaning representation \textit{mr}$_{src}$ to enhance the performance of SP models that convert \textit{u}$_{tgt}$ into \textit{mr}$_{tgt}$.\footnote{As shown in Figure \ref{fig:overview}, the meaning representation is largely language-independent, similar to Python code or SQL grammar, except for its slot values. Therefore, synthetic utterances in the target language that contain slot values from the source language are still useful for training SP models in target languages.}
CBP consists of two components: (1) an utterance generator (Section \ref{question_generator}) that synthesizes utterances in the target languages from source \textit{mr}; (2) a filtering mechanism (Section \ref{filter}) that discards low-quality synthesized utterances. To train the models in each step, we utilize SP datasets in the source language and monolingual corpora in both the source and target languages.

The utterance generator, which utilizes a seq2seq Transformer \cite{vaswani2017attention} as its backbone, is trained to generate \textit{u}$_{src}$ from the input \textit{mr}$_{src}$, and subsequently generates \textit{u}$_{tgt}$ from \textit{mr}$_{src}$ during inference.
To achieve this, the model must be capable of generating utterances in different languages from the same meaning representations.
Therefore, we introduce \emph{a language identity switch operation} and \emph{a language-specific adapter} to control the language of the generated utterances.

The language identity switch operation alters the encoder output representation of the generator to reflect the source language identity, ensuring that the generator’s decoder always receives the encoder representation with the source language identity regardless of the input language. We then train the utterance generator to produce output sequences in the target language using the modified encoder representation, while integrating language-specific adapters \cite{houlsby2019parameter} into the Transformer decoder. This training enables the adapter to prompt the generator to produce utterances in different languages while maintaining the same meaning from a given representation.

Then, we remove low-quality data from the synthesized utterances using the filtering mechanism. By re-parsing the generated utterances, we measure round-trip consistency \cite{alberti2019synthetic} to determine whether it accurately maps back to the input meaning representation used during generation. This data filtration process improves the quality of the synthesized data.

\subsection{\label{question_generator} Utterance generator}
\paragraph{Architecture} We construct the utterance generator using a multilingual pretrained seq2seq model, such as mT5~\cite{xue2021mt5}, as its backbone. To control the output languages, we integrate a language-specific adapter into each decoder block of the generator, positioning it immediately after the feed-forward layers. These adapters are lightweight bottleneck feed-forward layers that enable the generator to adapt to specific languages by learning modular representations \citep{pfeiffer2020mad, parovic2022bad}.
\paragraph{Training language adapters} As illustrated in Figure \ref{fig:step1}, we initially train the language adapters using monolingual corpora for each language, respectively.
Each language adapter is updated through a denoising task, where the utterance generator reconstructs randomly masked sentences into their original forms.
During this training process, the model learns solely from data where the input and output sequences share the same language.
However, during the data synthesis step, the model is required to generate an output sequence in the target language (\textit{u}$_{tgt}$) when provided with an input sequence in the source language (\textit{mr}$_{src}$). When we train the adapter with a conventional denoising objective \cite{lewis2020bart, ustun2021multilingual}, this mismatch leads to failure in synthesizing utterances in target languages (Figure~\ref{fig:mschema_gen_rate}). To mitigate this language mismatch in the zero-resource setting without parallel corpora, we propose a novel \textbf{source-switched denoising objective} to train the adapters, leveraging the representation geometry of mPLMs.

Previous studies~\cite{libovicky2020language,yang2021simple} have shown that the representation of mPLMs can be decomposed into language-specific and language-neutral components, which respectively capture language identity and semantic information.
Inspired by this property, we \textbf{switch} the language identity of input sequences to the source language during the denoising task to prevent the model from determining the output language based on the input language.
Following \citet{libovicky2020language}, we estimate the language-specific component for language $l$ as the language mean vector $\mu_{l}$. 
We compute $\mu_{l}$ as the mean of 1M contextualized token representations obtained from the encoder of the utterance generator, using a set of sentences from the monolingual corpora $C_l$.

\begin{figure}[!t]
    \centering
    \includegraphics[width=\columnwidth]{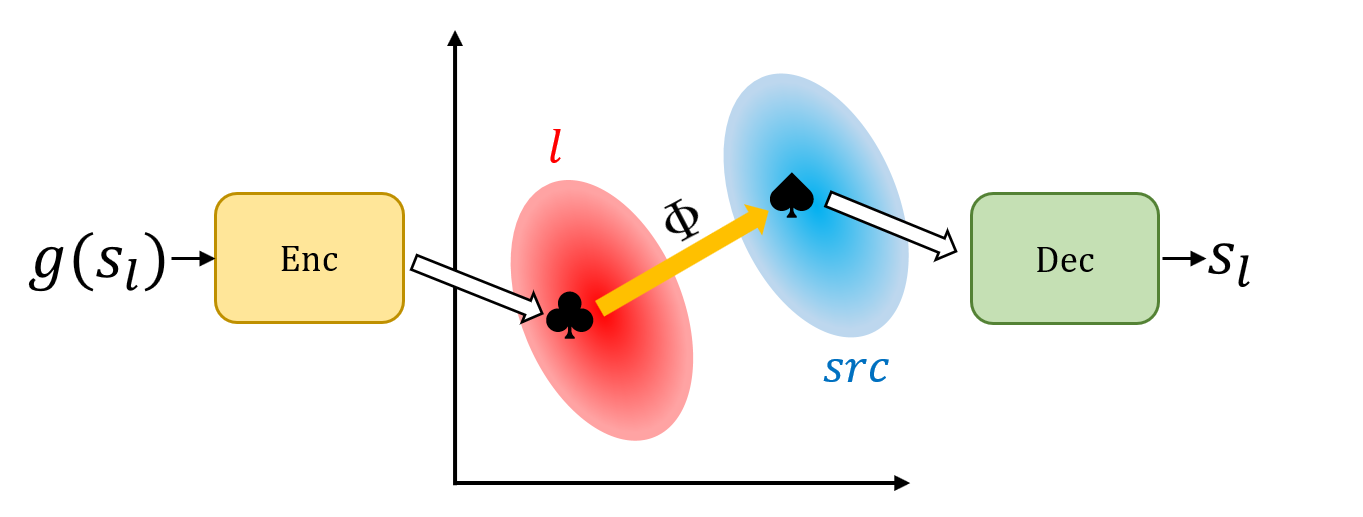}
    \caption{During source-switched denoising training, the language identity switch operation $\Phi$ switches the language identity of the encoded representation of the masked sentence \(\clubsuit\) from language $l$ to the source language, resulting in \(\spadesuit\).}
\label{fig:source_switched}
\end{figure}

During the training of the language adapter $A_l$, a masked sentence $g(s_l)$ in language $l$ is fed into the encoder $Enc$ of the utterance generator and encoded into a representation. 
We then modify the language-specific component of the encoded representation to the source language using the language identity switch operation $\Phi$. 
Formally, the operation is defined as:
\begin{eqnarray*}
\Phi(Enc(g(s_l))) = Enc(g(s_l)) - \mu_{l} + \mu_{src}
\end{eqnarray*}
where $\mu_{src}$ is a language mean vector of the source language. 
This operation maintains the semantic equivalence of the representation while changing its identity to the source language (Figure \ref{fig:source_switched}).

The language-specific adapter learns to map input sentences from the source language to sentences in each target language while preserving the meaning, using the source-switched denoising objective.
Initially, sentences are distorted using a noise function $g$, which replaces consecutive spans of the input sentence with a mask token.
The decoder then reconstructs the original sentence based on the encoder representation with the language identity switched to the source language.
For each language $l$, language adapter $A_l$ is separately trained to minimize $L_{A_l}$:
\begin{eqnarray*}
L_{A_{l}} = \sum_{s_l \in C_l}-log P(s_l|\Phi(Enc(g(s_l))); A_{l})
\end{eqnarray*}
where $s_l$ is a sentence belonging to monolingual corpora $C_l$ of language $l$. All utterance generator parameters are frozen during the training except those of the adapter. 

While we focus on training adapters in this work, these source-switched denoising training strategies can potentially be applied to other modular methods such as LoRA~\cite{hu2022lora}. We chose to focus on adapters for two main reasons: (1) they generally show better performance compared to other modular methods given the same size of trainable parameters~\cite{he2022towards}, and (2) the literature background on their usage for cross-lingual transfer~\cite{pfeiffer2020mad, pfeiffer-etal-2023-mmt5}.

\paragraph{Fine-tuning Utterance generator} After training language adapters for each language, we fine-tune the utterance generator to synthesize \textit{u}$_{src}$ from \textit{mr}$_{src}$ using labeled data in the source language, as shown in Figure~\ref{fig:step2}. This process involves integrating the \textbf{source} language adapter into the decoder and selectively freezing other layers of the utterance generator and the adapter to prevent catastrophic forgetting.\footnote{Inspired by~\citet{pfeiffer-etal-2023-mmt5}, we explore various freezing configurations to optimize utterance synthesis of the target language. Table \ref{tab:frozen_ablation} in the Appendix illustrates that the best results are obtained by additionally freezing the embedding layer, decoder attention, and cross-attention.}

\paragraph{Synthesizing target utterances} After fine-tuning the utterance generator, we synthesize \textit{u}$_{tgt}$ from \textit{mr}$_{src}$.
For each \textit{mr}$_{src}$ in the labeled data, we generate \textit{u}$_{tgt}$ across various target languages by incorporating the corresponding language-specific adapter $A_{tgt}$ into the decoder.

\subsection{Filtering mechanism} \label{filter}
To filter out low-quality synthesized utterances, we propose a filtering mechanism inspired by round-trip consistency~\cite{alberti2019synthetic}.
We fine-tune the same backbone model for the utterance generator for the SP task using only labeled data from the source language.
For each target language utterance \textit{u}$_{tgt}$ initially generated from \textit{mr}$_{src}$, the trained SP model predicts its corresponding meaning representation \textit{mr}$_{pred}$.
We use the set (\textit{u}$_{tgt}$, \textit{mr}$_{src}$) where \textit{mr}$_{src}$ exactly matches \textit{mr}$_{pred}$ to ensure that the synthesized \textit{u}$_{tgt}$ preserves the meaning of the \textit{mr}$_{src}$.

\section{Experimental Settings}
\subsection{Datasets}
To evaluate whether our methodology generalizes across different languages and meaning representations, we assess our methods on two cross-lingual SP datasets: Mschema2QA~\cite{zhang2023xsemplr} and Xspider~\cite{shi2022xricl}. Examples of each dataset are presented in Table~\ref{tab:dataset}.
\paragraph{Mschema2QA} is a question-answering dataset over schema.org web data that pairs user utterances with meaning representations in the ThingTalk Query Language. The dataset contains 8,932 training and 971 test examples, each available in 11 languages. Using English as the source language, we evaluate our model on the test split across 10 target languages\footnote{Arabic (ar), German (de), Spanish (es), Persian (fa), Finnish (fi), Italian (it), Japanese (ja), Polish (pl), Turkish (tr), and Chinese (zh)}. 
\paragraph{Xspider} is a cross-domain text-to-SQL dataset that pairs user utterances with SQL queries. We train our model on the English Spider dataset\cite{yu2018spider} consisting of 7,000 training examples and evaluate on the Chinese~\cite{min2019pilot} and Vietnamese~\cite{vitext2sql} dev split. We did not assess Farsi and Hindi as they are not publicly available. 

\begin{table}[!ht]
\centering
\small
    \begin{tabularx}{\columnwidth}{l|X}
        \hline
        {\color[HTML]{FE0000} \textit{u}} & quali sono i luoghi da \textbf{piazza barberini, 9} \\
        {\color[HTML]{FE0000} \textit{mr}} & now => ( @org.schema.Hotel.Hotel ) filter param:geo:Location == location: "\textbf{piazza barberini, 9}" => notify \\
        \hline
        {\color[HTML]{32CB00} \textit{u}} & \begin{CJK*}{UTF8}{bsmi}那里有多少员工？\end{CJK*} \\
        {\color[HTML]{32CB00} \textit{mr}} & SELECT count(*) FROM employee \\
        \hline
    \end{tabularx}
\caption{Italian and Chinese examples of utterances (\textit{u}) and corresponding meaning representations (\textit{mr}) for Mschema2QA (\citet{zhang2023xsemplr}, in \textcolor{red}{red}) and Xspider (\citet{shi2022xricl}, in \textcolor{green}{green}), respectively. Mschema2QA tends to have phrase-level slot values (in bold).}

\label{tab:dataset}
\end{table}

\paragraph{Monolingual corpora} We create unlabeled monolingual corpora $C_l$ for each language $l$ by extracting 1 million sentences from the November 20, 2023, Wikipedia dump in the respective language. We extract the raw article texts from the dump using WikiExtractor~\cite{Wikiextractor2015} and split them into sentences using BlingFire~\cite{blingfire}. 

\subsection{Implementation details}
We use the multilingual pretrained seq2seq model mT5-large~\cite{xue2021mt5} as the backbone for our SP model and utterance generator. The synthesized datasets for Mschema2QA and Xspider contain 49.4k and 8.2k examples, respectively. We train the model in a single stage using these synthesized datasets along with the labeled data in the source language ($D_{src}$), which is English. Employing AdamW~\cite{loshchilov2017decoupled} optimizer, we train the SP model for 50 epochs on both datasets, with a batch size of 32 and a learning rate of 3e-5. Appendix~\ref{sec:exp_setup} has further details.

\subsection{Baselines}
As the datasets have been proposed recently, few prior results are available in the literature. Therefore, we developed several strong baselines that do not use labeled datasets in target languages. All baselines, except those using LLM, utilize mT5-large as the backbone model.
\paragraph{Translation-Based Baselines}  For \textbf{Translate-Test}, we use Google Translate~\cite{wu2016google} to convert the target language test set into English and then input it into the model trained only with $D_{src}$. In \textbf{Translate-Train}, $D_{src}$ is translated into all target languages using machine translation (MT), and a model is trained on this data. For \textbf{TAP-Train}, we translate utterances from $D_{src}$ into all target languages with MT. Then, we use representative neural word aligners - awesome-align~\cite{dou2021word} - to align utterances with values from meaning representations, constructing a dataset to train a multilingual parser. In \textbf{TAP-Train + source}, we supplement the dataset from TAP-Train with $D_{src}$ to train the model.
\paragraph{In-Context Learning with Multilingual LLMs} We use \textbf{gpt-3.5-turbo}\footnote{https://platform.openai.com/docs/models/gpt-3-5-turbo} for in-context learning. The prompt is constructed by appending English examples and an utterance from the evaluation dataset, with eight examples for Mschema2QA and one for Xspider to meet input limits. For Xspider, we additionally compare against the state-of-the-art method that uses LLM, \textbf{DE-R$^{2}$+Translation-P}~\cite{shi2022xricl}.
\paragraph{Zero-Resource Baselines} For \textbf{Zero-shot}, we train a model using the English-labeled dataset $D_{src}$ only. In \textbf{word translation}, inspired by ~\citet{zhou-etal-2021-improving}, we create an augmented dataset by replacing words in English utterances from $D_{src}$ with their counterparts in the target language, using bilingual dictionaries from MUSE~\cite{conneau2017word}. To preserve alignment between the meaning representation and the utterance, we only replace words that are not part of the values. Models are trained using both $D_{src}$ and the word-replaced dataset across target languages. For \textbf{reconstruction}, inspired by~\citet{maurya-etal-2021-zmbart}, we train an SP model with an auxiliary task of reconstructing input from noisy data using unlabeled corpora across target languages. This reconstruction objective aims to enrich the cross-lingual latent representation space across languages.

Additionally, We report supervised performance as an upper bound, trained on data from all languages. We train baselines utilizing mT5 with the same hyperparameters and setup as the proposed method. Additional details for baseline models can be found in Appendix \ref{sec:exp_setup_baseline}.
\subsection{Evaluation metrics} We measure Exact Match (EM) accuracy for the Mschema2QA and XSpider datasets. Additionally, we report Test-suite (TS) accuracy for the XSpider dataset following ~\citet{zhong2020semantic}. Each score is averaged over three runs with different random seeds.
\section{Results and Analysis}

\begin{table*}[!hbt]
\centering
\resizebox{0.9\textwidth}{!}{
\begin{tabular}{lccccccccccc}
\Xhline{0.9pt}
\textbf{Model}      & ar            & de            & es            & fa            & fi            & it            & ja            & pl            & tr            & zh            & \textbf{Avg}  \\ \Xhline{0.9pt}
Supervised          & 52.8          & 68.1          & 68.9          & 46.7          & 65.9          & 63.5          & 61.9          & 59.0          & 63.7          & 53.7          & 60.4          \\ \Xhline{0.9pt}
Translate-Test      & 20.9          & 43.3          & 34.2          & 22.0          & 29.9          & 36.6          & 20.3          & 38.1          & 30.9          & 21.0          & 29.7          \\
Translate-Train     & 13.8          & 40.8          & 37.0          & 17.8          & 32.0          & 33.4          & 2.9           & 37.5          & 31.9          & 7.4           & 25.5          \\
TAP-Train           & 26.4          & 51.1          & 47.0          & 28.7          & 53.8          & 43.4          & 2.6           & 46.6          & 50.4          & 8.8           & 35.8          \\
TAP-Train + source  & 28.8          & 60.5          & 53.2          & 29.5          & 54.4          & 48.6          & 5.2           & 47.9          & 55.6          & 8.2           & 39.2          \\ \Xhline{0.9pt}
gpt-3.5-turbo             & 13.0          & 15.7          & 15.3          & 11.2          & 16.4          & 15.0          & 7.5           & 15.6          & 14.1          & 12.5          & 13.6          \\ \Xhline{0.9pt}
Zero-shot & \textbf{32.3} & 58.8          & 56.3          & 34.9          & 49.9          & 58.1          & 11.3          & 49.9          & 48.2          & 26.0 & 42.6          \\
{+ word translation}               & 30.3          & 63.4          & 60.3          & 31.1          & 50.3          & 60.3          & \textbf{13.7} & 52.9          & 52.9          & 21.1          & 43.7          \\
{+ reconstruction}           & 29.9          & 56.3          & 54.9          & 26.8          & 44.5          & 57.4          & 6.7           & 49.6          & 44.6          & \textbf{26.2}          & 39.7          \\ 
\textbf{CBP}            & 32.0          & \textbf{63.8} & \textbf{60.4} & \textbf{35.2} & \textbf{56.8} & \textbf{62.4} & 11.2          & \textbf{54.8} & \textbf{57.0} & 24.3          & \textbf{45.8} \\ \hdashline
\hspace{0.5em} {w\slash o filtering}            & 9.4          & 58.4 & 51.3 & 17.1 & 38.0 & 54.6 & 5.5 & 51.3 & 43.9 & 11.0          & 34.0 \\ 

\Xhline{0.9pt}
\end{tabular}
}
\caption{Exact match accuracy on Mschema2QA across (i) supervised models, (ii) translation-based models, (iii) LLM-based models, and (iv) zero-resource models. The best results among the zero-resource models are highlighted in bold.}

\label{tab:mschema2qa_res}
\end{table*}

\begin{table}[]
\centering
\resizebox{\columnwidth}{!}{%
\begin{tabular}{l|c|cc|cc}
\Xhline{0.9pt}
\multirow{2}{*}{\textbf{Model}} & zh-full & \multicolumn{2}{c|}{zh} & \multicolumn{2}{c}{vi} \\ \cline{2-6} 
 & EM & \multicolumn{1}{c|}{EM} & TS & \multicolumn{1}{c|}{EM} & TS \\ \Xhline{0.9pt}
Supervised & 61.3 & \multicolumn{1}{c|}{65.9} & 72.2 & \multicolumn{1}{c|}{57.4} & 58.5 \\ \Xhline{0.9pt}
Translate-Test & 56.5 & \multicolumn{1}{c|}{62.8} & 69.2 & \multicolumn{1}{c|}{35.2} & 38.3 \\
Translate-Train & 57.6 & \multicolumn{1}{c|}{63.5} & 70.6 & \multicolumn{1}{c|}{47.9} & 50.0 \\
TAP-Train & 59.6 & \multicolumn{1}{c|}{65.3} & 71.0 & \multicolumn{1}{c|}{54.2} & 50.1 \\
TAP-Train + source & 59.3 & \multicolumn{1}{c|}{64.9} & 69.8 & \multicolumn{1}{c|}{53.2} & 50.7 \\ \Xhline{0.9pt}
gpt-3.5-turbo & 38.0 & \multicolumn{1}{c|}{35.3} & 67.9 & \multicolumn{1}{c|}{37.5} & 55.0 \\
{DE-R$^{2}$+Translation-P$^\dag$} & 47.4 & \multicolumn{1}{c|}{52.7} & 55.7 & \multicolumn{1}{c|}{43.7} & 43.6 \\ \Xhline{0.9pt}
{Zero-shot}  & 43.7 & \multicolumn{1}{c|}{49.3} & 55.7 & \multicolumn{1}{c|}{47.4} & 46.9 \\
{+ word translation} & 41.0 & \multicolumn{1}{c|}{47.1} & 54.3 & \multicolumn{1}{c|}{47.0} & 45.3 \\
{+ reconstruction} & 42.8 & \multicolumn{1}{c|}{47.9} & 54.3 & \multicolumn{1}{c|}{\textbf{48.6}} & \textbf{47.5} \\ 
{\textbf{CBP}} & \textbf{47.8} & \multicolumn{1}{c|}{\textbf{54.0}} & \textbf{59.5} & \multicolumn{1}{c|}{47.3} & \textbf{47.5} \\ \hdashline
\hspace{0.5em} {w\slash o filtering} & 41.9 & \multicolumn{1}{c|}{47.5} & 54.1 & \multicolumn{1}{c|}{41.0} & 42.5 \\ \Xhline{0.9pt}
\end{tabular}%
}
\caption{Performance on Xspider. As only a subset of data in Cspider can be evaluated with TS, we reported zh and zh-full individually, following ~\citet{shi2022xricl}. † is taken from \citet{shi2022xricl}. The best results among the zero-resource models are highlighted in bold.}
\label{tab:xspider_result}
\end{table}

In Tables \ref{tab:mschema2qa_res} and \ref{tab:xspider_result}, we compare the performance of CBP against competitive baselines on the Mschema2QA and Xspider benchmarks. CBP improves the average EM score on Mschema2QA by 3.2\%, with significant improvements of 8.8\% in Turkish and 5.0\% in German, compared to the zero-shot method without data augmentation. Similarly, on Xspider, our method enhances Chinese performance by 4.7\% in EM and 3.8\% in TS. The filtering mechanism proves essential for our method, as evidenced by the significant drop in performance in its ablation (w/o filtering). Remarkably, despite operating under the zero-resource setting, our method outperforms all baseline models on the Mschema2QA dataset and even surpasses DE-R$^2$+Translation-P, the state-of-the-art in the literature on the Xspider dataset. These results highlight the effectiveness and practicality of CBP in cross-lingual SP.

Additionally, we find that gpt-3.5-turbo exhibits different performance trends on the two datasets. On Mschema2QA, gpt-3.5-turbo performs poorly, indicating that in-context learning with English examples alone is insufficient to learn the dataset’s domain-specific grammar. This highlights the practicality of zero-shot cross-lingual transfer through fine-tuning.
Conversely, on Xspider, where the model has pre-trained knowledge about text-to-SQL~\cite{liu2023comprehensive}, gpt-3.5-turbo shows strong performance, surpassing ours in TS. However, our backbone model, mT5-large (1.2B parameters), is notably more parameter-efficient and cost-effective than gpt-3.5-turbo.

\paragraph{Slot value alignment} One key challenge in cross-lingual data augmentation for SP is aligning slot values between the utterance and the meaning representation. Compared to translation-based baselines, we measure the slot value alignment rate of augmented data synthesized by CBP. The alignment rate is the percentage of training examples where the utterance contains exactly every slot value from the corresponding meaning representation.
Table~\ref{tab:value_alignment_rate} presents each dataset's average slot value alignment rates across languages.\footnote{Refer to Appendix \ref{sec:slot_appendix} for the process of extracting slot values and results across all target languages.} Notably, CBP consistently exhibits a \textbf{high alignment rate} across both the Xspider and Mschema2QA datasets. On the Mschema2QA dataset, our method achieves a slot alignment rate of 97.91\%, significantly higher than the 75.77\% achieved by the translate-and-project (TAP-Train) approach while performing competitively in Xspider. 

\begin{table}[h!]
\centering
\resizebox{\columnwidth}{!}{%
\begin{tabular}{l S[table-format=2.2] S[table-format=2.2] S[table-format=2.2]}
\toprule
\textbf{Dataset} & \textbf{Translate-Train} & \textbf{TAP-Train} & \textbf{CBP} \\
\midrule
Mschema2QA & 55.04 & 75.77 & \textbf{97.91} \\
Xspider    & 78.89 & \textbf{96.10} & 94.67 \\
\bottomrule
\end{tabular}
}
\caption{Slot value alignment rates of augmented datasets across various methods}
\label{tab:value_alignment_rate}
\end{table}

This discrepancy can be attributed to differences in average slot value length and complexity. In Xspider, the average slot value length is 7.75 characters and 1.31 words, whereas, in Mschema2QA, it is 15.38 characters and 2.38 words, making slot alignment with word aligner (TAP-Train) more challenging. Our method, however, maintains a high slot value alignment rate in Mschema2QA, demonstrating its effectiveness in cross-lingual SP tasks with longer slot values.

\begin{figure}[h]
    \centering
    \includegraphics[width=\columnwidth]{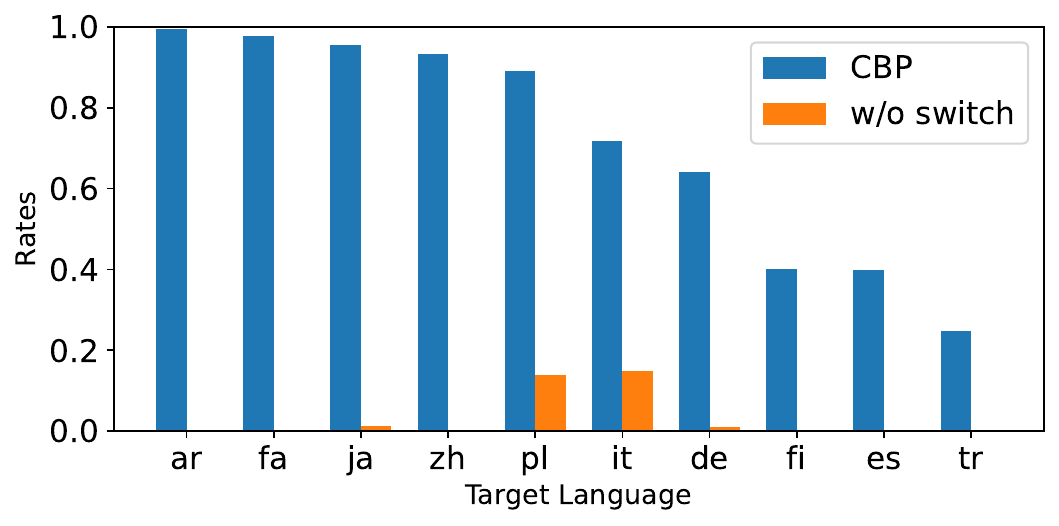}
    \caption{Target language synthesis rates (from 0 to 1) on different languages in Mschema2QA. For more granular results, refer to Appendix Table~\ref{tab:syn_dist}.}
    \label{fig:mschema_gen_rate}
\end{figure}

\paragraph{Target language synthesis rate}
\label{sec:target_synthesis_rate}
To evaluate the impact of the source-switched denoising training on synthesizing target language utterances, we assess the language of synthesized utterances in Mschema2QA using the Google Cloud Translation API’s Language Detection. Figure~\ref{fig:mschema_gen_rate} shows the target language synthesis rate. Training the language adapter with a conventional denoising objective (w/o switch) fails to synthesize target language utterances effectively. In contrast, our method, which employs a source-switched denoising objective, achieves a high synthesis rate in the target language, demonstrating its effectiveness.

Notably, our approach excels in synthesizing languages that do not share a script with the source language (English; Latin script), achieving high synthesis rates. Our method also performs robustly for languages sharing scripts, though at slightly lower rates. We speculate that shared scripts may result in similar language identities during language adapter training, potentially reducing differentiation between languages. Nevertheless, considering the widespread use of non-Latin scripts globally, CBP's consistent target synthesis rate with these scripts highlights its broad applicability and effectiveness.

\begin{figure}[h]
    \centering
    \includegraphics[width=0.9\columnwidth]{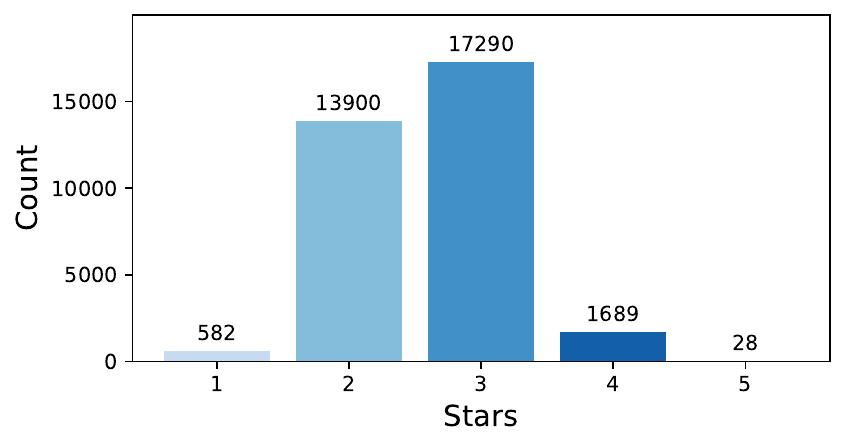}
    \caption{Quality of synthesized utterances measured by GEMBA-stars. We use gpt-3.5-turbo as the backbone.}
    \label{fig:quality}
\end{figure}

\paragraph{Quality of synthesized utterances} We evaluate the translation quality between the synthesized utterance from CBP and the English utterance paired with the meaning representation for the synthesized utterance. We assess the quality only for the synthesized utterances that were identified as being in the target language by the language detection API. We employ GEMBA-stars~\cite{kocmi2023large}, a state-of-the-art GPT-based metric that assesses translation quality on a one-to-five-star scale through zero-shot prompting. Figure~\ref{fig:quality} shows the star distribution for synthesized utterances across all languages on Mschema2QA. We find that the majority of utterances fall within the two to four-star range, indicating similar meaning to some degree. This suggests that our method not only adjusts the synthesized utterances' language but also preserves their meaning to some extent. Synthesized utterances across different languages are presented in Figure~\ref{fig:syn_examples} in the Appendix.

\paragraph{Monolingual data size}

\begin{figure}[h]
    \centering
    \includegraphics[width=0.8\columnwidth]{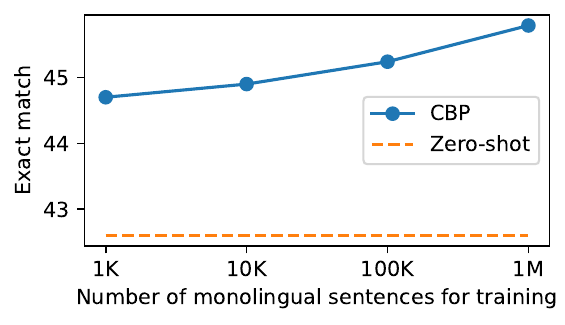}
    \caption{Average exact match on Mschema2QA across different languages with varying monolingual corpora sizes (1K to 1M)}
    \label{fig:mono_size}
\end{figure}

To assess the impact of monolingual corpora size in source-switched denoising training, we trained adapters for the languages used in Mschema2QA with progressively smaller data sizes. Figure~\ref{fig:mono_size} illustrates the average performance of Mschema2QA when trained with data sizes ranging from 1K to 1M across different languages. The results reveal better performance with more data, but notably, CBP outperformed the zero-shot method even with just 1K corpora per language. This demonstrates that our approach can be effective even for languages where acquiring large monolingual corpora is challenging.

\begin{figure}[h]
    \centering
    \includegraphics[width=\columnwidth]{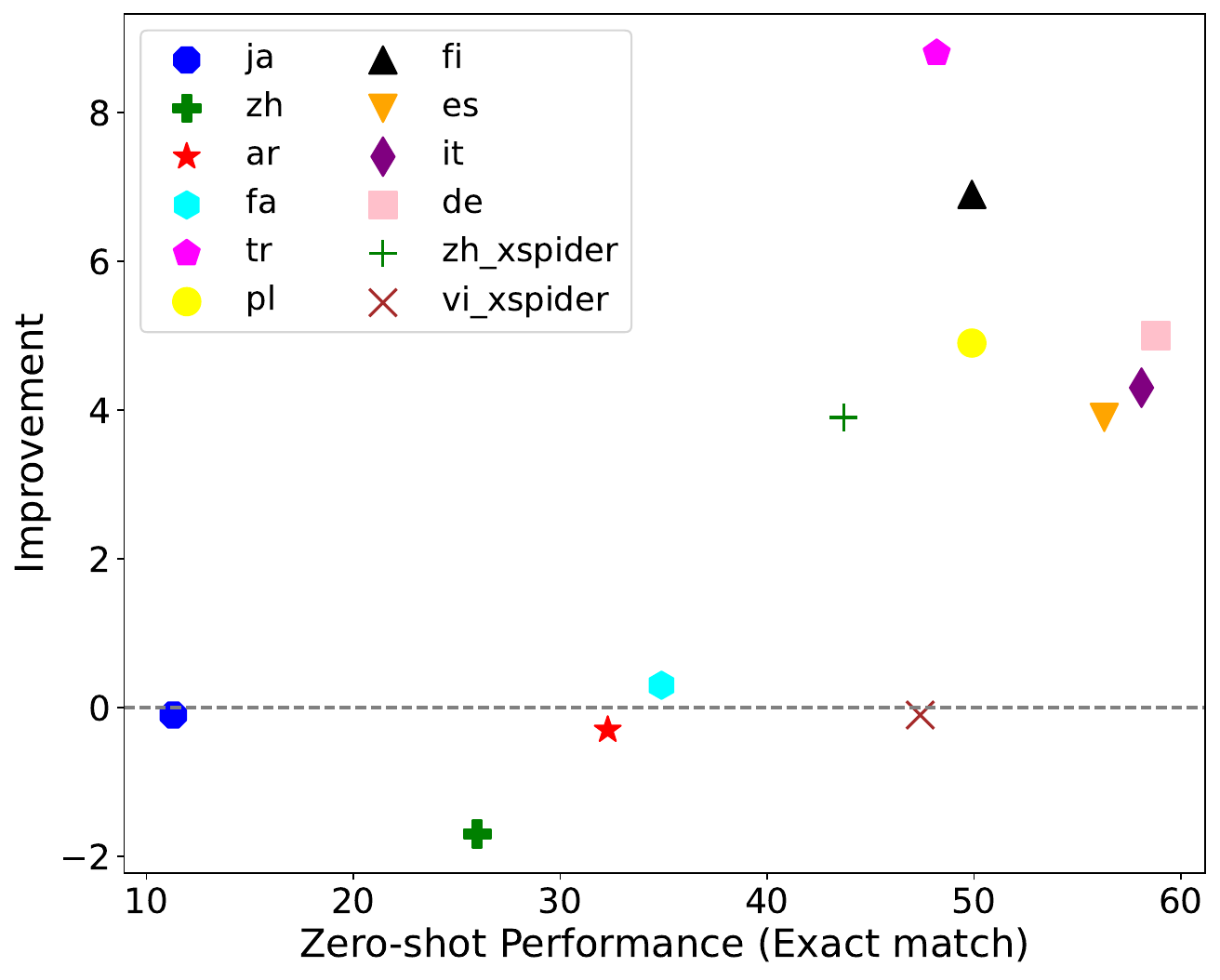}
    \caption{Relationship between zero-shot exact match performance and improvement through data augmentation (in exact match) across various languages. For those without the suffix "xspider," it pertains to mschema2qa.}
    \label{fig:gain_performance}
\end{figure}

\paragraph{Impact of zero-shot SP performance}
Figure~\ref{fig:gain_performance} illustrates the relationship between zero-shot EM performance and improvement through our data augmentation across various languages. The results show that languages with higher zero-shot performance tend to exhibit greater improvements from data augmentation. 
A notable comparison can be made between zh and zh\_xspider. Despite both being in Chinese, zh\_xspider, which has higher zero-shot task performance than zh, demonstrates significant gains from data augmentation, whereas zh does not. This indicates that the zero-shot performance of the task, rather than the language itself, is the primary factor influencing the magnitude of improvement through data augmentation. We hypothesize this is due to our method's use of zero-shot cross-lingual transferability of mPLMs for data synthesis and filtering, making initial zero-shot performance crucial. Future research could focus on enhancing data augmentation techniques to work effectively for languages with low zero-shot task performance.

\section{Conclusion}
We present Cross-lingual Back-Parsing (CBP), a novel data augmentation methodology aimed at enhancing zero-shot cross-lingual transfer for semantic parsing. Leveraging the representation geometry of multilingual pretrained language models, our method enables data augmentation in zero-resource settings. Our experiments on two cross-lingual semantic parsing benchmarks demonstrate that CBP significantly improves performance, underscoring its effectiveness and practical applicability. While we focus on semantic parsing, we believe that CBP has the potential to be applied to other cross-lingual generation tasks in zero-resource settings. Future work will investigate the application of our method to tasks such as cross-lingual text style transfer~\cite{krishna2022few}.

\section{Limitations}
Our proposed methodology, CBP, synthesizes target language utterances from source meaning representations by leveraging the representation geometry of mPLMs. Although we have demonstrated that CBP can effectively synthesize target utterances while preserving semantics, our experiments were conducted using only one mPLM (mT5-large). Validating our methodology with mPLMs of different parameter sizes and pretraining objectives would further demonstrate its generalizability. Additionally, while we demonstrated that our approach is beneficial even when the available monolingual corpora are small in size (Figure \ref{fig:mono_size}; applicable to actual low-resource language settings), we couldn't experiment on actual low-resource languages due to the limited natural language coverage of current semantic parsing datasets~\cite{zhang2023xsemplr}. Evaluating our methodology on actual low-resource languages could further verify its effectiveness. Finally, our methodology is less effective in synthesizing data when the zero-shot task performance is low. This indicates that our approach may not be effective for mPLMs with lower inherent performance, such as small-sized models. Future work could focus on improving our methodology to enhance performance even in these challenging scenarios.

\section*{Acknowledgements}
This research was supported by the MSIT(Ministry of Science and ICT), Korea, under the ITRC(Information Technology Research Center) support program(IITP-2024-2020-0-01789) supervised by the IITP(Institute for Information \& Communications Technology Planning \& Evaluation), and by Hyundai Mobis. We thank the anonymous reviewers for their helpful comments and suggestions.

\bibliography{custom}

\appendix

\section{Appendix}
\subsection{Implementation details}\label{sec:exp_setup}
\paragraph{Language adapter} We initialize the language adapter as a bottleneck feed-forward layer, following the configuration from ~\cite{pfeiffer2020mad}. We train the adapter individually on source-switched denoising objectives using 1M sentences extracted from Wikipedia in each target language. We train the model for 100K steps with a batch size of 32 and a learning rate of 1e-4. We use AdamW~\cite{loshchilov2017decoupled} as an optimizer and minimize cross-entropy loss between the predicted sentence and the ground truth sentence. We use span masking as the noise function g, following the pretraining approach of mBART~\cite{liu2020multilingual}. A span of tokens is substituted with the mask token, with its length randomly sampled by a Poisson distribution with lambda=3.5. The training process took 26 hours on four A100-80GB GPUs.

\paragraph{Question generator} We initialize the question generator with mT5-large~\cite{xue2021mt5}. We train the model for 50 epochs on both Mschema2QA and Xspider, with a batch size of 4 and a learning rate of 3e-5. We use AdamW~\cite{loshchilov2017decoupled} as an optimizer, and we minimize cross-entropy loss between the predicted question and the ground truth question. We used the final checkpoint to minimize errors during the utterance generation process. For Xspider, we append linearized databases to the input, following the style used by~\citet{li2023resdsql}. The training process took 8 hours for Xspider and 6 hours for Mschema2QA with four A100-80GB GPUs.
\paragraph{Semantic parser} We initialize the semantic parser with mT5-large~\cite{xue2021mt5}. We train the model for 50 epochs on both Mschema2QA and Xspider, with a batch size of 32 and a learning rate of 3e-5. We use AdamW~\cite{loshchilov2017decoupled} as an optimizer, and minimize cross-entropy loss between the predicted meaning representation and the ground truth representation. We select checkpoints based on validation performance on the English dev set. For Xspider, we append linearized databases to the input, following the style used by~\citet{li2023resdsql}. The training process took 5 hours for Xspider and 11 hours for MschemaQA with four RTX6000ADA GPUs. 

\subsection{Additional details for baseline models}\label{sec:exp_setup_baseline}
\begin{figure}[!h]
\centering
\resizebox{\columnwidth}{!}{
\includegraphics{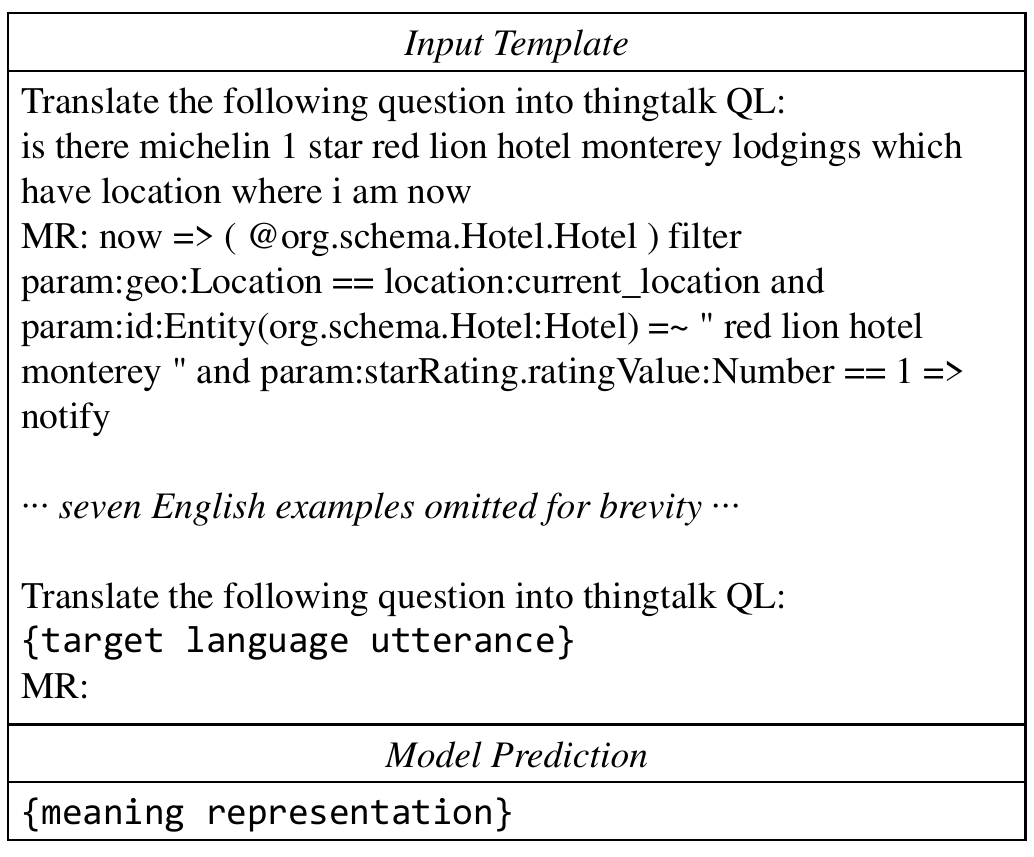}}
\caption{\label{fig:appendix:prompt_mqa} The input and output template for few-shot inference of GPT-3.5-turbo for Mschema2QA}
\end{figure}

\begin{figure}[!h]
\centering
\resizebox{\columnwidth}{!}{
\includegraphics{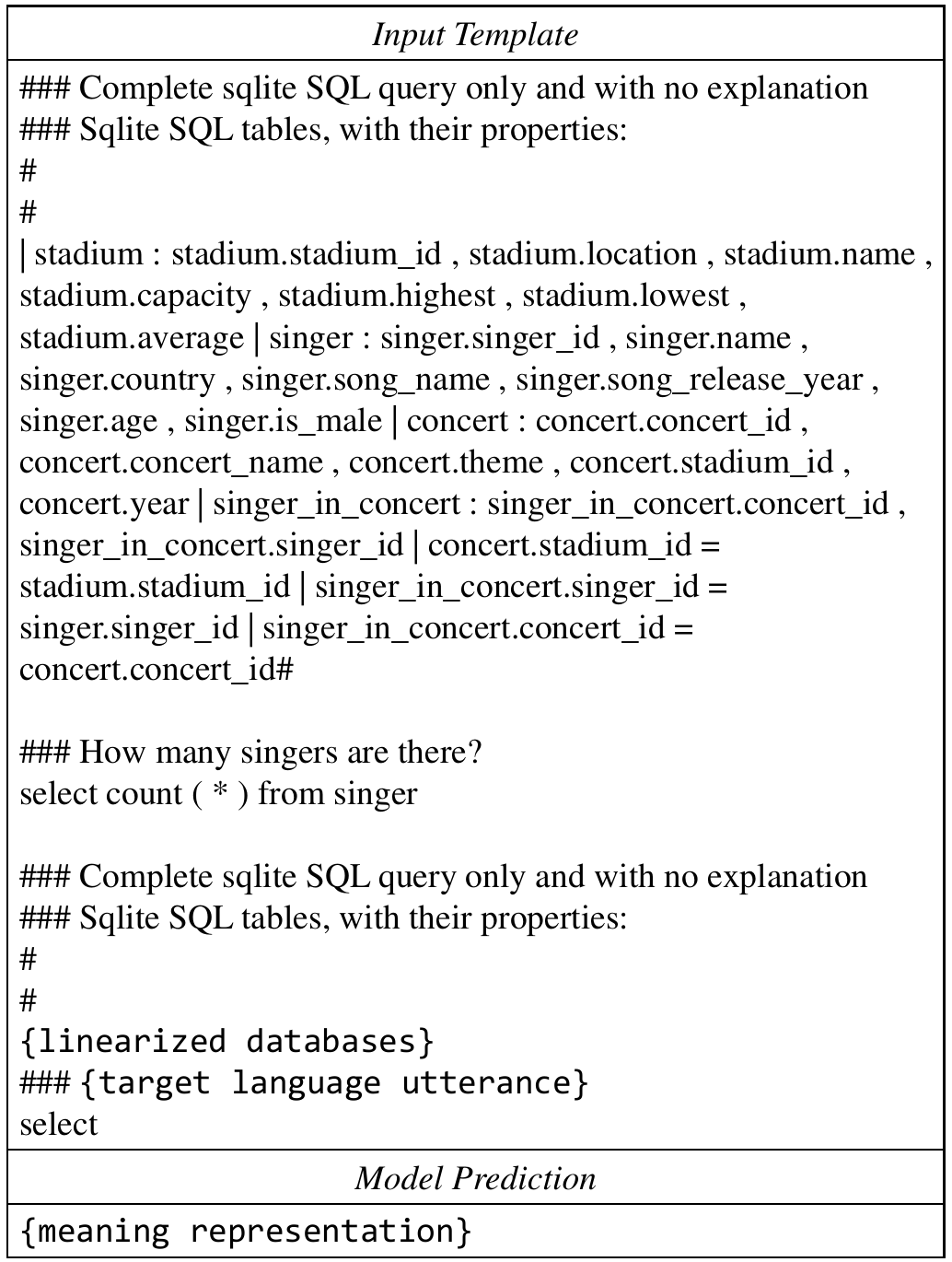}}
\caption{\label{fig:appendix:prompt_xspider} The input and output template for few-shot inference of GPT-3.5-turbo for Xspider. We used the prompt format in ~\cite{liu2023comprehensive}, while linearizing databases following styles used by ~\cite{li2023resdsql}. 
}
\end{figure}

In this section, we provide additional details for the following baselines: TAP-Train, gpt-3.5-turbo, and reconstruction.
\paragraph{TAP-Train} Both Mschema2QA and Xspider have slot values in their meaning representations enclosed in double quotes. We used regex to extract these slot values. Using the unsupervised version of awesome-align~\cite{dou2021word}, we replaced the slot values in the meaning representation with the corresponding values in the utterance. We found that replacing values in the utterance with the corresponding slot values from the meaning representation performed worse, so we opted to replace the slot values in the meaning representation instead.
\paragraph{GPT-3.5-turbo} We construct the prompt for gpt-3.5-turbo by appending English examples and an utterance in the target language from the evaluation dataset. To meet input limits, we include eight examples for Mschema2QA and one for Xspider, selected randomly. Figures~\ref{fig:appendix:prompt_mqa} and ~\ref{fig:appendix:prompt_xspider} show the prompt format for Mschema2QA and Xspider, respectively. We used regular expressions to post-process the model's predictions, extracting only the required meaning representations.

\paragraph{Reconstruction}
In addition to the cross-entropy loss $L_{SP}$ used for semantic parsing training, we introduce a loss $L_{RE}$ for the auxiliary task of reconstructing input from noisy data using unlabeled corpora across target languages. We utilized the same unlabeled corpora (Wikipedia) that were employed to train the language adapter. After extracting sentences from these corpora, we applied the identical noise function used in the language adapter training, which masks spans of tokens. The auxiliary task aims to reconstruct the original input from this noised input, and we utilized the cross-entropy loss between the predicted input and the original input for the task. The final loss is computed as follows:
\begin{eqnarray*}
L = L_{SP}+\alpha L_{RE},
\end{eqnarray*}
where $\alpha$ is the weight for $L_{RE}$. We empirically optimized $\alpha$ to 0.01 among candidates of [0.001, 0.01, 0.1, 0.5], as it performed the best in the evaluation.
\subsection{Slot value alignment results across languages}
\label{sec:slot_appendix}
Mschema2QA and Xspider include slot values in their meaning representations, enclosed in double quotes. We used regex to extract these slot values. We measure the slot value alignment rate as the percentage of examples where the utterance contains every slot value (EM) in the corresponding meaning representation. In cases where there are no slot values, we consider the alignment to be satisfied. To assess the alignment rate's impact on the model's performance with the augmented dataset, we compute the alignment rate across examples in the training set for each dataset. Table~\ref{tab:slot_language} presents the slot value alignment rates across languages in Mschema2QA and Xspider.

\begin{table*}[h!]
\centering
\resizebox{\textwidth}{!}{%
\begin{tabular}{lcccccccccc|cc}
\toprule
 & \multicolumn{10}{c|}{Mschema2QA} & \multicolumn{2}{c}{Xspider} \\
 & ar & de & es & fa & fi & it & ja & pl & tr & zh & zh & vi \\ 
\midrule
Translate-Train & 44.02 & 67.28 & 63.95 & 41.49 & 61.79 & 68.06 & 38.98 & 58.32 & 58.80 & 47.69 & 76.04 & 81.74 \\
TAP-Train       & 72.76 & 80.59 & 77.79 & 70.09 & 81.73 & 80.5 & 68.18 & 76.57 & 85.17 & 64.33 & \textbf{94.56} & \textbf{97.63} \\
\textbf{CPB}            & \textbf{98.69} & \textbf{98.78} & \textbf{99.03} & \textbf{97.75} & \textbf{99.21} & \textbf{99.09} & \textbf{92.16} & \textbf{98.74} & \textbf{98.9} & \textbf{96.76} & 93.15 & 96.18 \\ 
\bottomrule
\end{tabular}%
}
\caption{Slot value alignment results across languages. The best results are in bold.}
\label{tab:slot_language}
\end{table*}

\begin{table*}[!hbt]
\resizebox{\textwidth}{!}{%
\begin{tabular}{ccccccc|ccc}
\hline
\multicolumn{1}{l}{}                      & \multicolumn{1}{l}{}                         & \multicolumn{1}{l}{}                                    & \multicolumn{1}{l}{}                         & \multicolumn{1}{l}{}                          & \multicolumn{1}{l}{}                           & \multicolumn{1}{l|}{}                          & \multicolumn{3}{l}{Target synthesis rate} \\ \cline{8-10} 
\multicolumn{1}{l}{\multirow{-2}{*}{Emb}} & \multicolumn{1}{l}{\multirow{-2}{*}{Enc$_{LN}$}} & \multicolumn{1}{l}{\multirow{-2}{*}{Enc$_{Att}$ + Enc$_{FFN}$}} & \multicolumn{1}{l}{\multirow{-2}{*}{Dec$_{LN}$}} & \multicolumn{1}{l}{\multirow{-2}{*}{Dec$_{att}$}} & \multicolumn{1}{l}{\multirow{-2}{*}{Dec$_{catt}$}} & \multicolumn{1}{l|}{\multirow{-2}{*}{Dec$_{FFN}$}} & zh        & vi       & Avg.              \\ \hline
                                          &                                              &                                                         &                                              &                                               &                                                &                                                & 51.2      & 6.2      & 28.7              \\ 
X                                 & X                                    & X                                              &                                              &                                               &                                                &                                                & 50.5      & 13.9     & 32.2              \\ 
X                                 &                                              &                                                         & X                                    & X                                     & X                                      & X                                      & 93.3      & 15.6     & 54.4              \\ 
X                                 &                                              &                                                         &                                              & X                                    & X                                     &                                                & 88.0      & 27.7     & \textbf{57.8}     \\ 
X                                & X                                   &                                                         & X                                   & X                                    &                                                & X                                     & 75.0      & 11.6     & 43.3              \\ 
X                                & X                                   &                                                         & X                                   &                                               &                                                & X                                     & 84.9      & 6.8      & 45.8              \\ \hline
\end{tabular}%
}
\caption{Utterance synthesis rate in the target language of different freezing configurations, measured in Xspider. "X" denotes a frozen component.}
\label{tab:frozen_ablation}
\end{table*}

\begin{table*}[!t]
\resizebox{\textwidth}{!}{%
\begin{tabular}{ll|cccccccccc|cc}
\hline
\multirow{2}{*}{Method}    & \multirow{2}{*}{\begin{tabular}[c]{@{}l@{}}Synthesized\\ Language\end{tabular}} & \multicolumn{10}{c|}{Mschema2QA}                                              & \multicolumn{2}{c}{Xspider} \\ \cline{3-14} 
                           &                                                                                 & ar    & de    & es    & fa    & fi    & it    & ja    & pl    & tr    & zh    & zh           & vi           \\ \hline
\multirow{3}{*}{w/o switch} & target                                                                          & 0.00  & 0.00  & 1.18  & 0.00  & 13.99 & 14.18 & 0.92  & 0.00  & 0.00  & 0.00  & 0.02         & 10.65        \\
                           & source (en)                                                                     & 99.82 & 98.64 & 96.86 & 98.86 & 85.47 & 84.38 & 98.84 & 99.62 & 99.36 & 99.71 & 99.87        & 89.33        \\
                           & others                                                                          & 0.18  & 1.36  & 1.96  & 1.14  & 0.54  & 1.44  & 0.24  & 0.38  & 0.64  & 0.29  & 0.11         & 0.02         \\ \hline
\multirow{3}{*}{\textbf{CBP}}      & target                                                                          & 99.58 & 64.10 & 39.93 & 97.75 & 40.15 & 71.73 & 95.52 & 89.01 & 24.82 & 93.34 & 88.02        & 27.65        \\
                           & source (en)                                                                     & 0.30  & 35.42 & 57.82 & 1.93  & 59.17 & 26.32 & 3.84  & 10.88 & 74.32 & 6.23  & 11.37        & 72.35        \\
                           & others                                                                          & 0.12  & 0.48  & 2.25  & 0.32  & 0.68  & 1.95  & 0.64  & 0.11  & 0.86  & 0.43  & 0.61         & 0.00         \\ \hline

\end{tabular}}
\caption{Language distribution of synthesized utterances using language adapters trained with different methods}
\label{tab:syn_dist}
\end{table*}

\begin{figure*}[!h]
    \centering
    \includegraphics[width=2\columnwidth]{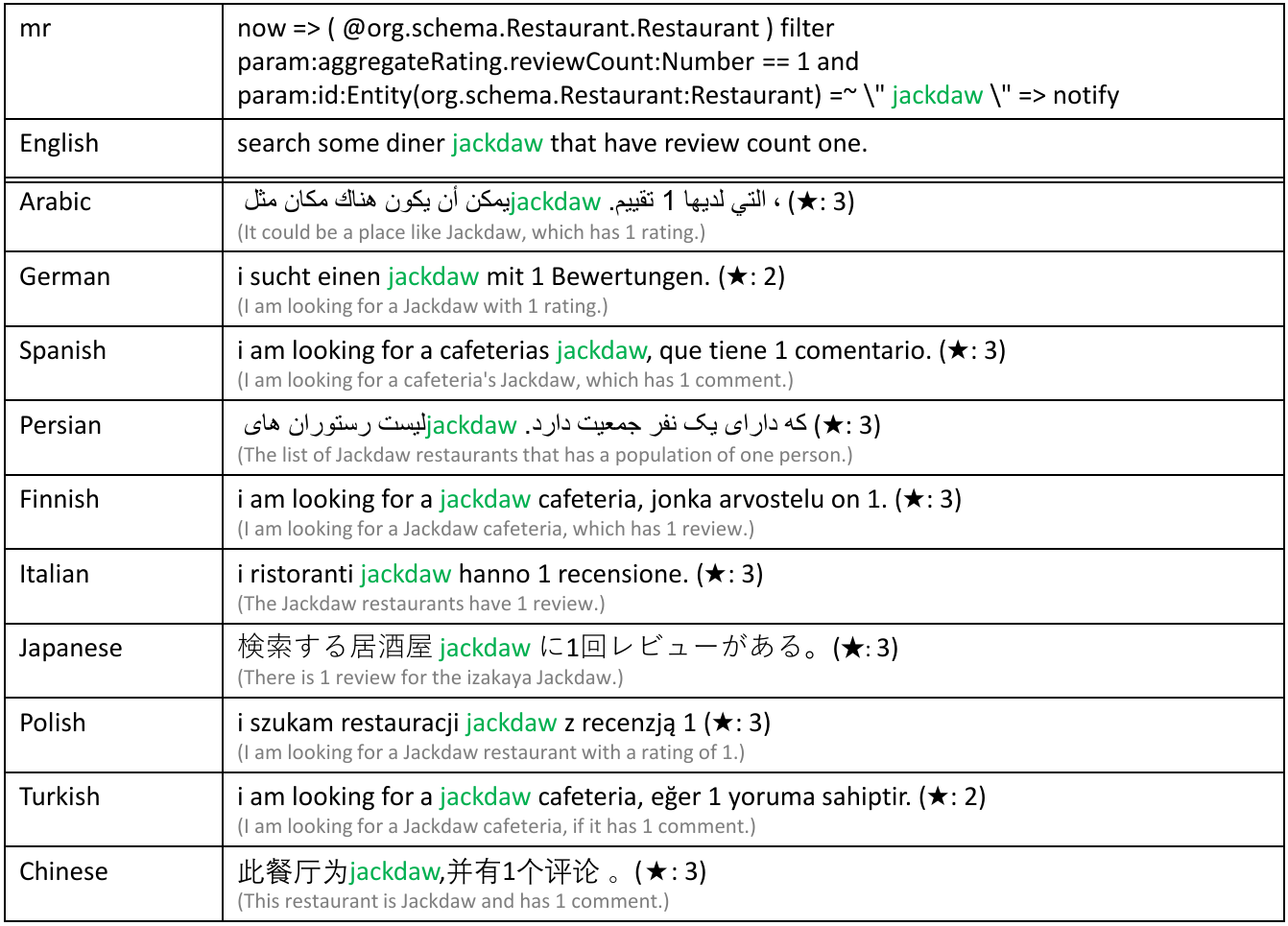}
    \caption{An example of synthesized utterances generated from a meaning representation of Mschema2QA~\cite{zhang2023xsemplr}. The synthesized utterances for each language are presented along with their corresponding English translations in parentheses. The numbers next to the stars indicate translation quality measured by GEMBA-stars~\cite{kocmi2023large}. The synthesized utterances convey a meaning similar to that of the English utterance. Additionally, the slot values remain unchanged in the synthesized utterances (in green).}
\label{fig:syn_examples}
\end{figure*}

\end{document}